\documentclass[conference]{IEEEtran}
\usepackage{amsmath,amssymb,amsfonts}
\usepackage{algorithmic}
\usepackage{graphicx}
\usepackage{textcomp}
\usepackage{xcolor}
\usepackage{booktabs}
\usepackage{csvsimple}
\usepackage{balance}

\usepackage[style=ieee,maxcitenames=1,mincitenames=1]{biblatex}
\usepackage{ifthen}
\makeatletter
\newcounter{IEEE@bibentries}
\renewcommand\IEEEtriggeratref[1]{%
  \renewbibmacro{finentry}{%
    \stepcounter{IEEE@bibentries}%
    \ifthenelse{\equal{\value{IEEE@bibentries}}{#1}}
    {\finentry\@IEEEtriggercmd}
    {\finentry}%
  }%
}
\makeatother

\begin{document}

\title{Estimates for the Branching Factors\\of Atari Games}

\author{\IEEEauthorblockN{Mark J.\ Nelson}
\IEEEauthorblockA{
\textit{American University}\\
Washington, DC, USA \\
mnelson@american.edu}
}

\maketitle

\begin{abstract}
The branching factor of a game is the average number of new states reachable
from a given state. It is a widely used metric in AI research on board games,
but less often computed or discussed for videogames. This paper provides
estimates for the branching factors of 103 Atari 2600 games, as implemented in
the Arcade Learning Environment (ALE). Depending on the game, ALE exposes
between 3 and 18 available actions per frame of gameplay, which is an upper
bound on branching factor. This paper shows, based on an enumeration of the
first 1 million distinct states reachable in each game, that the average
branching factor is usually much lower, in many games barely above 1. In
addition to reporting the branching factors, this paper aims to clarify what
constitutes a distinct state in ALE.
\end{abstract}

\begin{IEEEkeywords}
branching factor, Atari, Arcade Learning Environment
\end{IEEEkeywords}

\section{Introduction}

Atari 2600 games have been a popular challenge domain for artificial
intelligence research since the 2012 release of the Arcade Learning Environment
(ALE). ALE wraps the Atari emulator Stella in a framework familiar to AI
researchers: agents observe and take actions in an environment, sometimes
receiving positive or negative reward as a result \cite{ALE}. As of this
writing, ALE supports 104 games.

Each supported game has been manually instrumented by the ALE developers. For
example, ALE provides rewards to agents by reading changes in score from
game-specific locations in the Atari RAM. Part of this instrumentation is the
\emph{minimal action set}, the set of actions that have any effect in the game.
Currently supported games have minimal action sets as small as 3 and as large
as 18. For example, in \emph{Breakout} there are four actions: no-op (no
input), left, right, and fire. The maximum 18 are: no-op, the fire button, the
8 directions that can be registered by the Atari joystick, and each of those 8
directions while also pressing the fire button.

A game's minimal action set is an upper bound on its \emph{branching factor},
the number of new states that can be reached from a given state. Atari AI
research -- and other videogame AI research, for that matter -- does not normally
give much weight to (or even compute) branching factors. But they are commonly
discussed in AI board-game playing. The difference in branching factors is cited
as a reason that computer chess is harder than checkers, shogi harder than
chess, and go harder than the other three \cite{CheckersChess,Matsubara1996}.
Multi-game engines such as Ludii also compute branching factor as an
informational measure~\cite{LudiiManual:2021}. Does it give useful information
for video games? A first step in investigating that question is to compute
it.

A game's branching factor can be less -- often significantly less -- than its
minimal action set, for two reasons. The first is that the state space of many
games forms a graph, not a tree, so some distinct input sequences result in an
identical state. For example, in \emph{Tetris}, rotating a piece twice
clockwise will often result in the same state as rotating it twice
counter-clockwise. And in \emph{Breakout}, there are many ways the player can
move when the ball is in the air that will result in the same contact once the
ball comes back down.

The second reason that the average branching factor can be smaller than the
size of the action set is that Atari games may simply ignore some or all input
at various times. For example, once the player initiates a jump in
\emph{Q*bert}, all input is ignored for a number of frames while the jump
animation plays out.  And in \emph{Space Invaders}, the fire button has no
effect when the player already has a laser-cannon shot in the air, since the
game rules only allow one shot in the air at a time.

This paper provides average branching factor estimates for 103 of the 104 games
supported by ALE.\footnote{One is omitted because its initial state is broken;
see Section~\ref{sec:broken}.} These estimates are computed by exhaustively
enumerating the first 1 million distinct states reachable in each game. The
primary result is therefore Table~\ref{table:results}.

As a necessary prerequisite for enumerating those 1 million distinct states,
this paper also clarifies what constitutes a ``distinct state'' in ALE, in the
sense of \emph{state} meant by game-tree search and Markov decision processes
(MDPs). As discussed in Section~\ref{sec:state}, existing literature has
tended to either under- or over-specify state. In fact, a precise state for 6 of
the games (those that use the Atari paddle controller) is not retrievable from
the public API of ALE.

Finally, some lingering issues with determinism in ALE (a longstanding problem)
are uncovered by the experiments here. They only impact the branching factor
estimates of two games in a non-negligible way, but point to potential issues
with reproducibility.

\section{Methodology}

The main results of this paper are the estimated branching factors given in
Table~\ref{table:results}. This table is computed by enumerating the first 1
million distinct states in the games supported by ALE, and using those counts
to estimate branching factor.\footnote{More precisely, counting how many
distinct states are reachable in $n$ frames, until the first frame where $n
\geq 1,000,000$.} Carrying out this task requires: a definition of what
constitutes a distinct \emph{state}, a method for estimating \emph{branching
factor}, and a deterministic emulator. All three of these are surprisingly
tricky to pin down.

\subsection{Atari state}
\label{sec:state}

\begin{figure}
\includegraphics[width=0.24\textwidth]{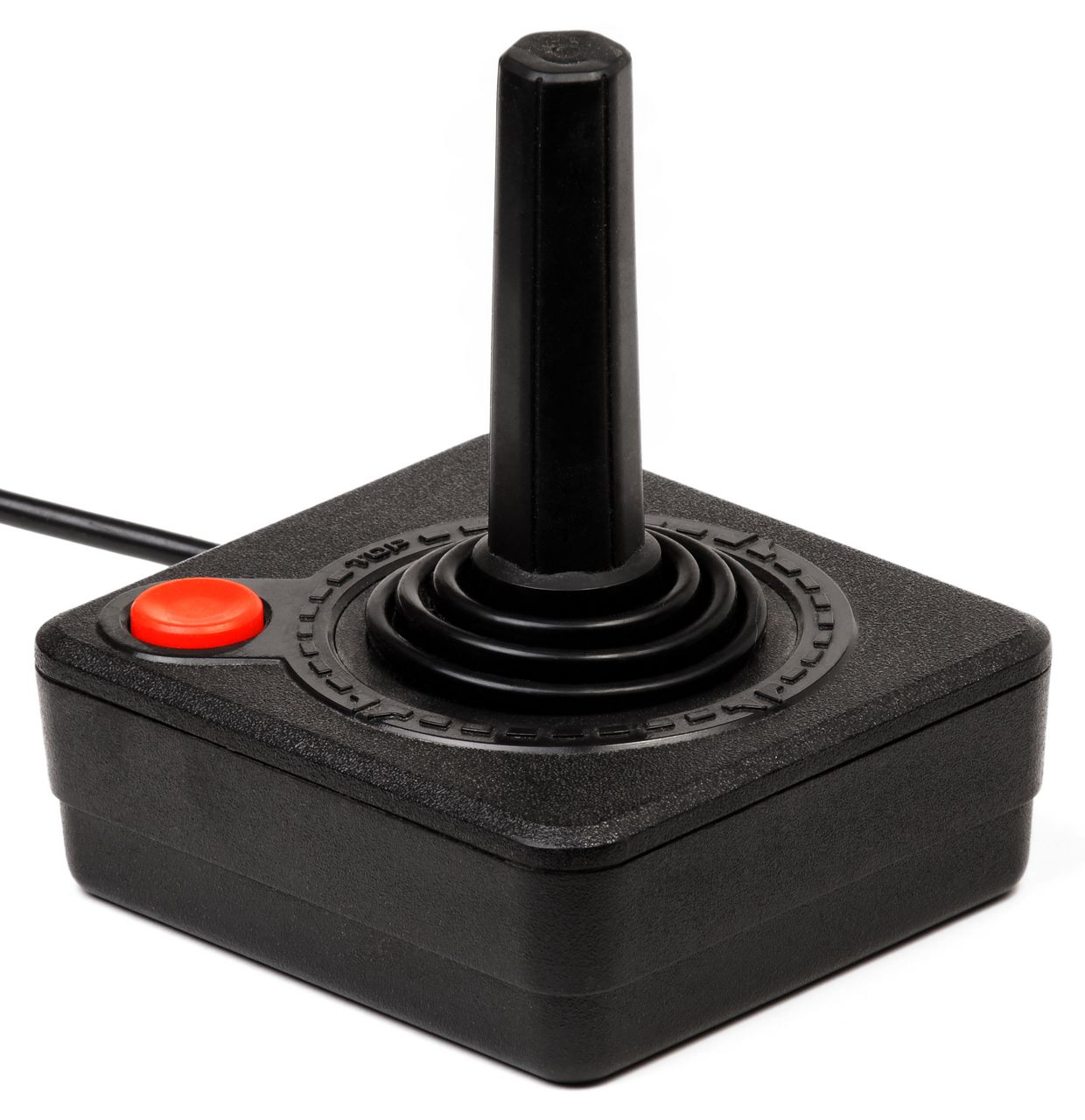}
\includegraphics[width=0.24\textwidth]{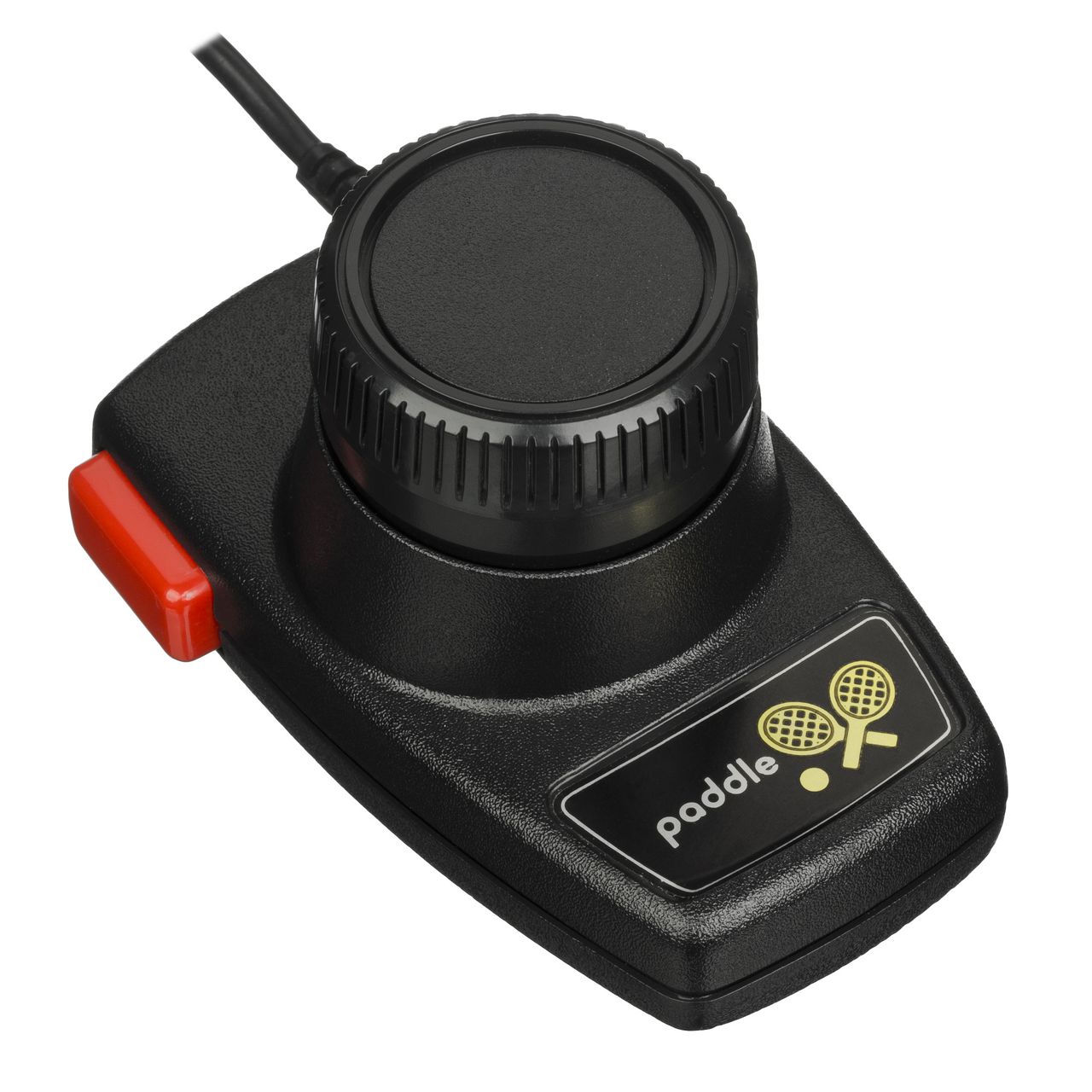}
\caption{The two Atari input devices emulated by ALE: Joystick on the left, and
paddle controller on the right. (Photographs by Evan Amos; released into the
public domain.)}
\label{fig:input}
\end{figure}

Tree search and reinforcement learning algorithms both have a concept of
\emph{state}. A state for such algorithms is enough information to uniquely
determine an environment's future dynamics. Given the same state, the same
action sequence will produce the same sequence of successor states if the
environment is deterministic; or it will produce the same distribution over
state sequences if the environment is stochastic.

For one-player Atari games, the state can be completely specified by the 128
bytes of Atari RAM, plus the current position of the paddle controller, if the
game uses the paddle controller. To the best of my knowledge, this paper is the
first to use this definition of an Atari game state, which I claim to be the
correct one. Computing it required patching ALE, since ALE doesn't expose
paddle position in the public API.\footnote{The patched version of ALE used in
the experiments reported in this paper, which also applies a determinism fix
discussed in Section~\ref{sec:determinism}, is available at
\url{https://github.com/NelsonAU/Arcade-Learning-Environment}.}

Previous definitions of Atari/ALE game state either under- or overspecify the
state:

\subsubsection{Underspecified game state}

Much existing literature assumes that the Atari RAM is sufficient to capture
the game state. For example, \textcite{ALE:revisiting} mention that ALE allows
querying either the current screen image or the current RAM, and call the RAM
the ``real state'':

\begin{quote}
This observation can be a single 210×160 image and/or the current 1024-bit RAM
state. Because a single image typically does not satisfy the Markov property,
we distinguish between observations and the environment state, with the RAM
data being the real state of the emulator.
\end{quote}

The RAM is in fact sufficient for many Atari games. The Atari 2600 has no
external storage, network interface, etc., so the only place it can store data
across frames is in RAM. However, it also polls every frame for player input
from a physical controller. And some Atari controllers -- one of which is used
by ALE -- maintain their own external state.

ALE supports two emulated player input devices: the joystick and the paddle
(Figure~\ref{fig:input}). The joystick does not maintain external state. It
returns one of the 18 possible actions each frame, with no interaction between
frames (at least if we assume a player with sufficiently fast hands, such as an
AI bot). Therefore the Atari RAM is sufficient to capure state when using the
joystick.

The paddle, though, does maintain external state. It is a rotating
potentiometer -- a wheel that sends different voltages to the console depending
on its position. Therefore, when playing on a real Atari console, the
current position of the paddle in the physical world is part of the game state.
The way ALE implements the paddle is that  the actions ``left'' and ``right''
increase or decrease the the paddle's current rotation by a compile-time
constant, \texttt{PADDLE\_DELTA}, up to specified maximum and minimum
values.\footnote{Paddle constants are defined in
\texttt{src/environment/ale\_state.hpp}.} Therefore in the case of ALE as well,
the paddle's rotation is needed in addition to the Atari RAM to have a complete
game state.

Of the 104 ALE-supported games, 98 use the joystick, and six use the paddle.
The six are: backgammon, blackjack, breakout, casino, kaboom, and pong. The
original ALE paper~\cite{ALE} makes a point of noting the added complexity
posed by paddle controllers as an example of how ALE captures some of the
messiness of real-world decision-making, but subsequent papers often ignore
them.

\subsubsection{Overspecified game state}

ALE also has a mechanism for serializing and deserializing state. This captures the
entire state of the emulator and ALE itself. But it is \emph{too} precise to
correspond to what AI algorithms normally mean by state, and is therefore also
not usable for this paper.

For example, serialized ALE states include the frame number, so if an otherwise
identical game state can be reached at frame 5, or through a different action
sequence at frame 6, using state serialization as the representation will treat
these as distinct states. That is not normal practice in AI algorithms.
Mechanisms such as discounted reward might prefer reaching the same state
sooner rather than later, but the time a state is reached is not part of the
\emph{definition} of a state.

\subsection{Branching factor}

The term \emph{branching factor} is used in a several different ways, all
relating to the number of successor states reachable from a given game state.
It can mean the \emph{maximum} branching factor, or the \emph{average}
branching factor. \textcite{Korf:1985} also distinguishes an \emph{edge
branching factor} (average number of outgoing legal moves) from
a \emph{node branching factor} (average number of new states reached through
such moves). The difference amounts to treating the game's state space as a
tree vs.\ graph: the node branching factor avoids double-counting states
reachable through multiple paths.

This paper estimates average node branching factors by a state-counting method.
First, exhaustively count the number of distinct states that can be reached,
through any sequence of actions, by a given frame number.\footnote{The
state-counting code used for this paper, along with data and analysis scripts,
is available in the \texttt{cog2021} branch of
\url{https://github.com/NelsonAU/ale_countstates/}} Then use this cumulative
count to estimate the branching factor as follows.

Observe that, if a game's average node branching factor were $b$, the
cumulative number of distinct game states $s$ observed by frame $f$ should be
approximately:

$$s \approx \sum_{i=0}^f b^i$$

This holds exactly if the game tree is uniform, because there will be exactly
$b^f$ new states reachable at frame $f$. Otherwise, it approaches
equality as $s$ grows large. We can rewrite this equation such that solving for
$b$ reduces to finding the (positive real) root of a polynomial:

$$(1-s) + b + b^2 + \ldots + b^f = 0$$

Such problems can be solved quickly to high precision by a number of
computational root-finding methods. The results in this paper use the
\texttt{uniroot} method of R v4.1.0.

\subsection{Determinism}
\label{sec:determinism}

\begin{table}[tbp]
\caption{Mismatches Between BFS and ID Estimates\\of Branching Factor (to 10k States)}
\begin{center}
\begin{filecontents*}{data/mismatches.csv}
Game,BFS,ID
pitfall2,5.04,6.48
space\_war,3.85,4.49
assault,4.51,4.55
solaris,7.61,7.64
ice\_hockey,1.09,1.09
frostbite,1.10,1.10
seaquest,1.05,1.05
tic\_tac\_toe\_3d,1.09,1.09
hangman,1.13,1.13
freeway,1.08,1.08
\end{filecontents*}
\csvautobooktabular{data/mismatches.csv}
\label{table:mismatches}
\end{center}
\end{table}

In principle, Atari games are deterministic
\cite{ALE:revisiting,ALE:determinism}.
In practice, ALE has often not been very deterministic,
due to a mixture of bugs and the complexity of emulation.
Bellemare notes: ``ALE determinism has always been brittle at
best''.\footnote{GitHub issue comment, January 12, 2020.
\url{https://github.com/mgbellemare/Arcade-Learning-Environment/issues/291#issuecomment-573483143}}

The patched version of ALE used in paper includes a significant patch from
Jesse Farebrother that improves ALE's
determinism.\footnote{\url{https://github.com/JesseFarebro/Arcade-Learning-Environment/tree/rng}}
It does solve most determinism problems, but a few games still behaved
strangely in preliminary experiments.

To test the reproducibility of estimating branching factor by exhaustively
counting distinct states, I compared the results of textbook versions of
iterative deepening (ID) and breadth first search (BFS) on a validation run
counting the first 10,000 distinct states. The two search methods agreed on 93
of 103 games. They disagreed on the 10 games shown in
Table~\ref{table:mismatches}, sorted from largest to smallest disagreement.

For understanding branching factor specifically, this table is perhaps not too 
concerning. Ninety-three games (those not in the table) have exact agreement
between ID and BFS. Of the ten games that differ, the branching factor
difference is below 0.1 in eight, and below 0.01 in six. Two do have noticeably
different estimates: \texttt{pitfall2} and \texttt{space\_war}.

The overall results for those two games are therefore worth taking with a grain
of salt. In addition, though it seemingly does not pose a large problem for
branching factor estimates specifically, the fact that iterative deepening and
breadth-first searches may produce different results on a deterministic domain
is somewhat worrying for the reproducibility of other analyses of the Atari
state space.

\subsection{Broken initial states}
\label{sec:broken}

Two games, \texttt{trondead} and \texttt{klax}, have an additional issue. Both
start in a kind of ``dead'' state: the initial RAM is not changed by any action
(neither game uses the paddle controller, so RAM is sufficient for state). That
would imply the games have only one total state, the initial one.  This seems
to be caused by the emulator not having properly initialized the RAM yet. The
game \texttt{trondead} was fixed by taking one no-op action before starting the
real experimment. The game \texttt{klax} seems to require more involved fix-up,
so was excluded.

\section{Results and discussion}

\begin{table*}[tbp]
\caption{Estimated Branching Factors (to 1mm states)}
\begin{center}
\begin{filecontents*}{data/results1.csv}
Game,Actions,Branching factor
haunted\_house,18,7.70
solaris,18,7.40
double\_dunk,18,7.15
zaxxon,18,5.01
boxing,18,3.92
assault,7,3.61
video\_pinball,9,3.42
yars\_revenge,18,3.17
road\_runner,18,3.17
laser\_gates,18,3.11
gravitar,18,3.11
darkchambers,18,3.10
sir\_lancelot,6,2.98
trondead,18,2.87
space\_war,18,2.86
enduro,9,2.66
word\_zapper,18,2.66
pitfall2,18,2.63
jamesbond,18,2.60
asteroids,14,2.42
robotank,18,2.32
private\_eye,18,2.32
tennis,18,2.18
centipede,18,2.14
earthworld,18,1.95
bank\_heist,18,1.88
basic\_math,6,1.88
air\_raid,6,1.81
backgammon,3,1.61
adventure,18,1.59
crossbow,18,1.59
entombed,18,1.58
skiing,3,1.58
berzerk,18,1.54
phoenix,8,1.49
asterix,9,1.48
montezuma\_revenge,18,1.45
donkey\_kong,18,1.41
carnival,6,1.41
pitfall,18,1.39
miniature\_golf,18,1.39
journey\_escape,16,1.38
human\_cannonball,18,1.35
elevator\_action,18,1.35
breakout,4,1.30
demon\_attack,6,1.30
alien,18,1.25
tutankham,8,1.24
fishing\_derby,18,1.24
keystone\_kapers,14,1.22
superman,18,1.20
atlantis2,4,1.19
\end{filecontents*}
\csvautobooktabular{data/results1.csv}
\begin{filecontents*}{data/results2.csv}
Game,Actions,Branching factor
atlantis,4,1.19
lost\_luggage,9,1.19
up\_n\_down,6,1.17
riverraid,18,1.16
chopper\_command,18,1.13
frostbite,18,1.12
mario\_bros,18,1.11
ice\_hockey,18,1.11
galaxian,6,1.11
venture,18,1.10
krull,18,1.10
videocube,18,1.10
hero,18,1.10
battle\_zone,18,1.09
surround,5,1.08
seaquest,18,1.08
kaboom,4,1.08
hangman,18,1.07
othello,10,1.07
pacman,5,1.07
space\_invaders,6,1.07
flag\_capture,18,1.07
tetris,5,1.07
defender,18,1.06
pong,6,1.06
videochess,10,1.06
time\_pilot,10,1.06
et,18,1.05
video\_checkers,5,1.04
turmoil,12,1.04
bowling,6,1.04
wizard\_of\_wor,10,1.04
kangaroo,18,1.04
star\_gunner,18,1.04
beam\_rider,9,1.04
pooyan,6,1.03
amidar,10,1.03
mr\_do,10,1.03
ms\_pacman,9,1.03
kung\_fu\_master,14,1.03
qbert,6,1.03
crazy\_climber,9,1.03
name\_this\_game,6,1.03
frogger,5,1.02
casino,4,1.02
blackjack,4,1.02
freeway,3,1.02
king\_kong,6,1.01
koolaid,9,1.01
gopher,8,1.01
tic\_tac\_toe\_3d,10,1.01
,,
\end{filecontents*}
\csvautobooktabular{data/results2.csv}
\label{table:results}
\end{center}
\end{table*}

This paper's main goal is to compute the branching factor estimates in
Table~\ref{table:results}. These results are computed by counting the distinct
states reachable in each game from the initial state, using breadth-first
search, until the frame at which the count exceeds 1 million. The experiments
took about 86.5 core-hours of CPU time on an Intel Xeon E5-2650 v4 @ 2.20GHz
(avg.\ of about 50 core-minutes per game)

The estimated branching factor is very low in a large number of games. The
median is 1.19, and 79 of 103 games (77\%) are below 2.0. And in a few dozen
games, the branching factors barely exceed 1.0.

Such low branching factors suggest that there may not be significant decisions
to be made every frame. An open question is what specifically that means.
Some existing work has proposed making decisions less often than every frame,
using a hyperparameter called frame-skip \cite{FrameSkip}. The results here
suggest frame-skip may be justified, but don't directly prove it. Future work
might look at whether branching factor correlates with optimal choice of
frame-skip. Alternately, if we first find the optimal frame-skip for a game, we
might recompute effective branching factors for the slower ``real'' decision
cycle.

A different explanation is that quick-reaction decision-making \emph{is}
sometimes needed, and so we can't simply use a more granular decision-making
cycle -- but only sometimes. The branching factors reported here are averages;
it is likely that some games have significantly more frame-to-frame variation
in the branching factor than others. Future work might look at additional ways
of summarizing the state space. Average branching factor is a single-number
summary of the rate of growth of a game's state space. Other measures might
look at variability or asymmetry of growth.

Finally, the relationship between branching factor and difficulty is open. In
board game AI, higher branching factor is often taken to mean a more complex
game (at least for AI!). Is this true for videogames?  My intention is to put
forward some numbers as a way of starting a discussion on that question.

\section*{Acknowledgment}

This work was supported by the National Science Foundation under Grant
IIS-1948017. Computing resources were provided by the American University High
Performance Computing System, which was funded in part by the National Science
Foundation under Grant BCS-1039497. 

Thanks to Amy Hoover, David Dunleavy, and the anonymous reviewers for helpful
suggestions.

\IEEEtriggeratref{4} 
\printbibliography

\end{document}